\documentclass[nohyperref]{article}

\usepackage{microtype}
\usepackage{graphicx}
\usepackage{subfigure}
\usepackage{booktabs} 

\usepackage{hyperref}

\usepackage[accepted]{icml2022}

\usepackage{amsmath}
\usepackage{amssymb}
\usepackage{mathtools}
\usepackage{amsthm}

\usepackage{mathrsfs}
\usepackage{multirow}
\usepackage{makecell}
\usepackage{color}
\usepackage{graphicx}
\usepackage{subfigure}
\usepackage{marvosym}
\usepackage[capitalize,noabbrev]{cleveref}

\theoremstyle{plain}

\theoremstyle{definition}

\theoremstyle{remark}

\usepackage[textsize=tiny]{todonotes}

\icmltitlerunning{Language Guided Image Inpainting with Defect-free VQGAN}

\begin{document}

\twocolumn[
\icmltitle{NÜWA-LIP: Language Guided Image Inpainting with Defect-free VQGAN}



\icmlsetsymbol{equal}{*}

\begin{icmlauthorlist}
\icmlauthor{Minheng Ni}{equal,hit}
\icmlauthor{Chenfei Wu}{equal,msra}
\icmlauthor{Haoyang Huang}{msra}
\icmlauthor{Daxin Jiang}{stca}
\icmlauthor{Wangmeng Zuo$\textsuperscript{\Letter}$}{hit}
\icmlauthor{Nan Duan$\textsuperscript{\Letter}$}{msra}
\end{icmlauthorlist}

\icmlaffiliation{hit}{Faculty of Computing, Harbin Institute of Technology, Harbin, China}
\icmlaffiliation{msra}{Microsoft Research Asia, Beijing, China}
\icmlaffiliation{stca}{Microsoft, Beijing, China}

\icmlcorrespondingauthor{Wangmeng Zuo}{wmzuo@hit.edu.cn}
\icmlcorrespondingauthor{Nan Duan}{nanduan@microsoft.com}

\icmlkeywords{Machine Learning, ICML}

\vskip 0.3in
]



\printAffiliationsAndNotice{\icmlEqualContribution} 

\begin{abstract}

Language guided image inpainting aims to fill in the defective regions of an image under the guidance of text while keeping non-defective regions unchanged. However, the encoding process of existing models suffers from either receptive spreading of defective regions or information loss of non-defective regions, giving rise to visually unappealing inpainting results. To address the above issues, this paper proposes NÜWA-LIP by incorporating defect-free VQGAN (DF-VQGAN) with multi-perspective sequence to sequence (MP-S2S). In particular, DF-VQGAN introduces relative estimation to control receptive spreading and adopts symmetrical connections to protect information. MP-S2S further enhances visual information from complementary perspectives, including both low-level pixels and high-level tokens. Experiments show that DF-VQGAN performs more robustness than VQGAN. To evaluate the inpainting performance of our model, we built up 3 open-domain benchmarks, where NÜWA-LIP is also superior to recent strong baselines.

\end{abstract}

\section{Introduction}



Image inpainting aims to fill missing pixels in the defective regions of a corrupted image such that the completed image is photo-realistic. It is a representative task in computer vision \cite{peng2021generating} \cite{zhao2020uctgan} \cite{zheng2019pluralistic} with many real-world practical applications, such as image manipulation, synthesis, object removal, etc.

With the rapid development of vision-language research, language-guided image inpainting becomes more and more an encouraging topic. Such systems perform inpainting using not only given pixels but also natural language instructions, and can be implemented based on GAN models \cite{bau2021paint} or diffusion models \cite{esser2021imagebart}. 
Recently, autoregressive model-based multimodal pre-training has exhibited strong visual synthesis capabilities, such as DALL-E \cite{ramesh2021zero}, CogView \cite{ding2021cogview} and NÜWA \cite{wu2021n}.
In particular, NÜWA has shown a good zero-shot text-guided image manipulation capability, opening the possibility of such a generative pre-training mechanism for language guided image inpainting.


However, existing pre-trained models suffer from the receptive spreading of defective regions, leading to inaccurate modeling of non-defective regions. Moreover, information loss of non-defective regions in autoregressive-based models, such as DALL-E, CogView, and NÜWA, results in non-trivial modification of this part. These two types of issues give rise to visually unappealing inpainting results.

To solve these problems, we propose NÜWA-LIP by leveraging a novel defect-free VQGAN (DF-VQGAN) and multi-perspective sequence to sequence (MP-S2S).
In contrast to VQGAN~\cite{esser2021taming}, DF-VQGAN incorporates the relative estimation for decoupling non-defective regions from defective regions to control receptive spreading. 
To protect the information of non-defective regions, symmetrical connections replenish the lost information of non-defective information from the encoding procedure.
Additionally, MP-S2S further enhances visual information from complementary perspectives, including both low-level pixels and high-level tokens under the text guidance.

Furthermore, we build three evaluation datasets for language-guided image inpainting and perform a comprehensive comparison with existing works.
Experiments show that NÜWA-LIP can achieve state-of-the-art results on these three benchmarks.
We also do ablation studies to reveal the effectiveness of different components in NÜWA-LIP.

To sum up, the major contributions of this paper are:
\begin{itemize}
    \item We propose a defect-free VQGAN, which introduces relative estimation to control receptive spreading and adopts symmetrical connections to protect information.
    \item We propose a multi-perspective sequence to sequence, which enhances visual information from complementary perspectives, including both low-level pixels and high-level tokens under the text guidance.
    \item We build three evaluation datasets for language guided image inpainting. Experiments show that NÜWA-LIP achieved state-of-the-art results compared to several strong baselines.
\end{itemize}

\section{Related Work}

\paragraph{Language-guided Image Inpainting}

Recent years have witnessed rapid progress in text-image synthesis that utilizes text to synthesize and manipulate images.
Language-guided image inpainting, as a subfield of text-image synthesis, has attracted more and more attention.
This task aims to fill in the defective regions of the image with the text guidance which describes the content of the full image.
Some works, such as \cite{bau2021paint} introduced generative adversarial networks to handle generic real images.
Several recent projects, such as ImageBART~\cite{esser2021imagebart}, Blended~\cite{avrahami2021blended}, and GLIDE~\cite{nichol2021glide}, perform this task with a diffusion model.
Following DALL-E~\cite{ramesh2021zero}, some works like NÜWA~\cite{wu2021n} utilize autoregressive model-based multimodal pre-training.
However, existing models suffer from either receptive spreading of defective regions or information loss of non-defective regions. Moreover, modeling the image in the one-fold way also limits the quality of the language guided image inpainting.

\paragraph{Vector Quantized Variational AutoEncoder}

Vector Quantized Variational AutoEncoder \cite{oord2017neural} is a VAE model which can compress the continuous information to discrete latent variables. Some works, such as VQVAE-2 \cite{razavi2019generating}, try to decode the image at a more fine-grained level. To decode a more vivid image, VQGAN \cite{esser2021taming} uses a GAN model to discriminate the generated image from the original image. However, the defective regions in the image will affect all discrete latent variables due to receptive spreading. This may lead to color cast or faults in the image. Meanwhile, the information loss will cause the modification of the non-defective region, thereby limiting the quality of the inpainting result.

\section{Method}

\begin{figure*}
	\centering
	\includegraphics[width=16cm]{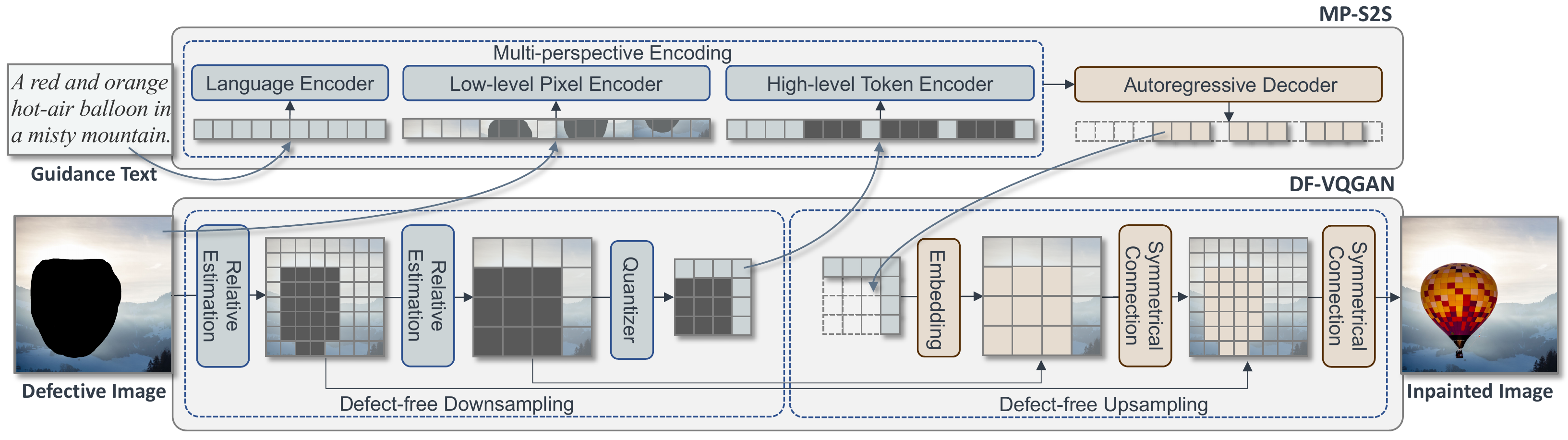}
	\caption{Overview of NÜWA-LIP.}
	\label{fig:model}
\end{figure*}
\subsection{Problem Formulation}

Given an input image $x\in \mathbb{R}^{W\times H\times C}$ with width $W$, height $H$ and channel $C$, a mask matrix $m\in \left\{0,1\right\}^{W\times H}$ with value $1$ denoting the defective regions, and a piece of natural text $t$, the language guided image inpainting aims to repair the defective regions under the guidance of the text and output a new image $y\in \mathbb{R}^{W\times H\times C}$. 
Hereafter, we call the input $x$ \textbf{defective image} and the output $y$ \textbf{inpainted image}.

In the Bayesian framework, the language-guided image inpainting task can be defined as maximizing the log posterior probability\footnote{We assume $t$ is conditionally independent of $y$ given $z, x, m$.}, as denoted in Eq.~(\ref{eq:p}):
\begin{equation} \label{eq:p}
\begin{aligned}
&\log p(y|x, m, t;\theta,\phi,\psi)\\
=&\log\frac{p(y|z,x,m;\psi)p(z|x,m,t;\theta)}{p(z|y,x,m;\omega)}
\end{aligned}
\end{equation}
where $\theta,\phi,\psi$ denote model parameters. $z$ denotes the latent variable. By taking the expectation w.r.t a auxiliary density $z\sim q(z|y,x,m;\phi)$ on both sides, the right side of Eq.~(\ref{eq:p}) can be rewritten as:
\begin{equation} \label{eq:e}
\begin{aligned}
&\mathbb{E}_{z\sim q(z|y,x,m;\phi)}[\log p(y|z, x, m;\psi)]\\
-&\mathbb{KL}_{z\sim q(z|y,x,m;\phi)}\left[q(z|y,x,m;\phi)||p(z|x, m, t;\theta)\right]\\
+&\mathbb{KL}_{z\sim q(z|y,x,m;\phi)}\left[q(z|y,x,m;\phi)||p(z|y, x, m;\omega)\right].
\end{aligned}
\end{equation}
According to VAE, since the third Kullback-Leibler divergence term is always greater than 0, we only need to maximize the first two terms denoted as the Evidence Lower BOund (ELBO). The first expectation term is the reconstruction loss of the inpainted image. The second term is a Kullback-Leibler divergence loss, which constrains the conditional distribution of the latent variable generated by the VAE decoder should be close to that generated by the auxiliary probability density. As Fig.~\ref{fig:model}, we will introduce how we model the first term with a defect-free VQGAN (DF-VQGAN) in Sec.~\ref{sec: df} and the second term with a multi-perspective sequence to sequence (MP-S2S) in Sec.~\ref{sec:mp}. 


\subsection{DF-VQGAN}\label{sec: df}

\begin{figure}[h]
	\centering
	\includegraphics[width=3.0in]{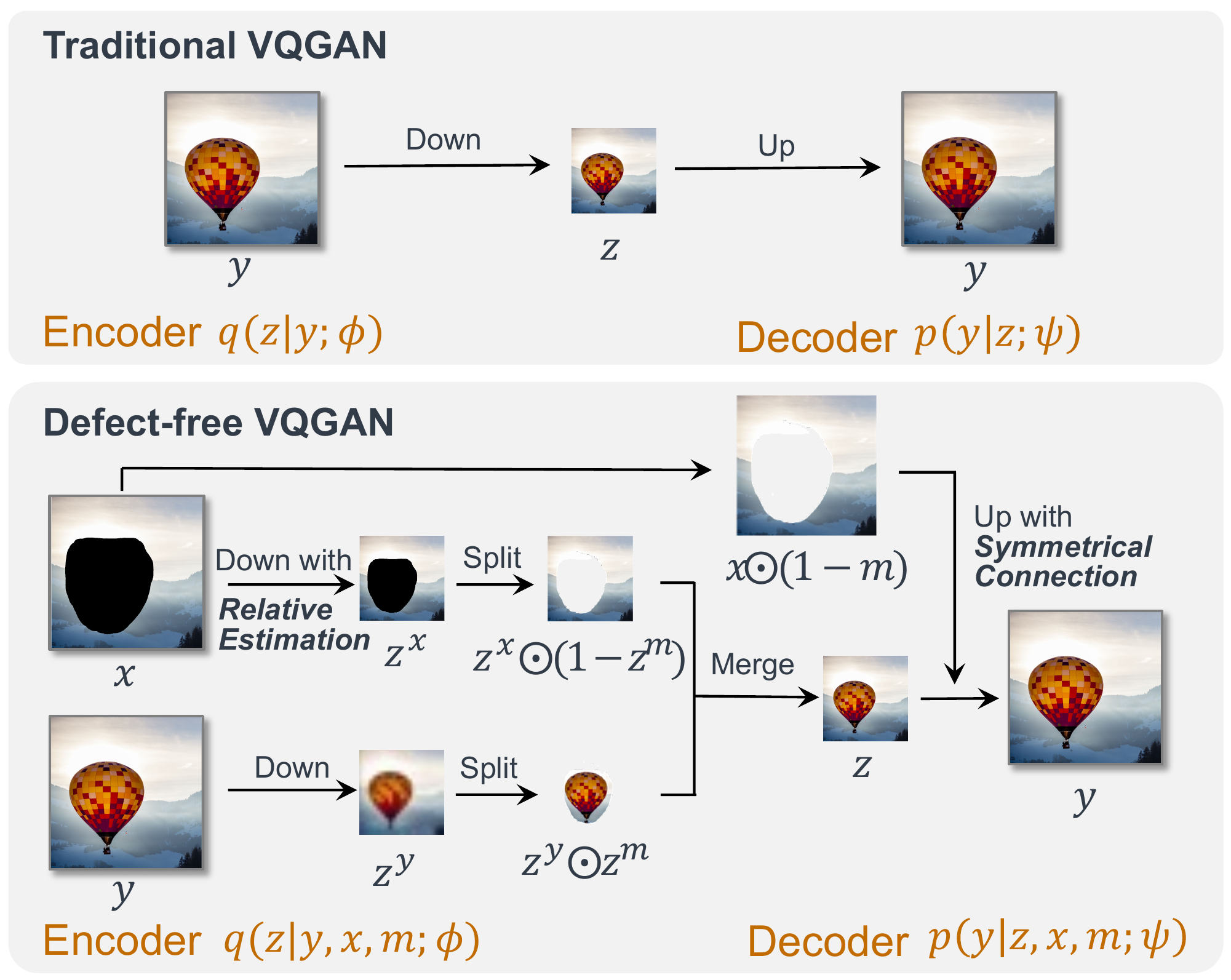}
	\caption{Comparisons between traditional VQGAN and our proposed DF-VQGAN during training. $x$ is the defective image, $y$ is the ground-truth inpainted image, $\hat{y}$ is the reconstructed inpainted image.  $m$ is a mask matrix. $z^x, z^y, z$ are latent variables. $z^m$ is the mask matrix which is max-pooled to the same shape of $z$. \textbf{\textit{relative estimation}} and \textbf{\textit{symmetrical connection}} are two crucial models to enable DF-VQGAN to generate non-defective $\hat{y}$.}
	\label{fig:df}
\end{figure}

The modeling of $\mathbb{E}_{z\sim q(z|y,x,m;\phi)}[\log p(y|z, x, m;\psi)]$ in Eq.~(\ref{eq:e}) can be split into a VAE Encoder of $q(z|y,x,m;\phi)$ and a VAE Decoder of $p(y|z, x, m;\psi)$. The probabilistic density function $q(z|y,x,m;\phi)$ shows that the probability distribution of $z$ should be conditioned given three variables: inpainted image $y$, defective image $x$ and mask matrix $m$. In other words, the latent variable $z$ should not only represent the inpainted image but also be sensitive to defective images with masked regions. To achieve this, we propose DF-VQGAN, a defective free VAE model based on the strong VQAGN backbone. To make it simple, we first introduce how to build a DF-VQGAN with a one-layer encoder and one-layer decoder in Fig.~\ref{fig:df}, and then introduce how to extend it to multiple layers.

\paragraph{Defect-free Downsampling} Firstly, the ground-truth inpainted image $y\in \mathbb{R}^{W\times H\times C}$ is fed into a VQGAN encoder layer, as shown in Eq.~(\ref{eq:zy}):
\begin{equation} \label{eq:zy}
z^y=\mathrm{e}(y),
\end{equation}
where $\mathrm{e}$ denotes a downsampling function, which typically consists of several normal operations, such as attentions or convolutions. $z^y\in \mathbb{R}^{w\times h\times d}$ denotes the encoding result. To encode the defective image $x$, we note that different from $y$, the defective regions (masked regions) in $x$ will affect non-defective regions during the downsampling (See Fig.~\ref{fig:df}). 


Could we only encode the non-defective regions in $x$ to prevent the effects from defective regions? Thanks to the mask matrix, we could easily mask the defective parts for attention and convolution operations. Using the re-estimated mean and variance, the normalization\footnote{To simplify the formulation, we use Layer Norm to reveal our method, but this equation can be extended easily to other normalization functions, such as Group Norm.} can be given by, 
\begin{equation} \label{eq:layernorm}
\mathrm{Norm}^{\mathrm{DF}}(x, m)=\frac{x-\frac{N}{N_m}\mathrm{E}[x]}{\sqrt{\frac{N-1}{N_m-1}\mathrm{Var}[x']}+\epsilon},
\end{equation}
where $N$ denotes the number of pixels in $x$, and $N_m$ denotes the number of defective pixels. $x'$ denotes the $x$ with the defective region fulfilled with $\frac{N}{N_m}\mathrm{E}[x]$. Then, we could replace the operations in Eq. (3) with a set of defect-free operations to get the defect-free encoder $\mathrm{e}^{\mathrm{DF}}$ and carefully encode the defective image $x\in \mathbb{R}^{W\times H\times C}$:
\begin{equation} \label{eq:zx}
z^x=\mathrm{e}^{\mathrm{DF}}(x, m),
\end{equation}
where $z^x\in \mathbb{R}^{w\times h\times d}$ is the down-sampling result without polluting the non-defective part.
We named this defect-free down sampling step with \textit{\textbf{relative estimation}}. Note that $\mathrm{e}^{\mathrm{DF}}$ shares the same parameter with the $\mathrm{e}$.

By a simple max pooling, the mask matrix can be resized to fit the size of $z^x$ and $z^y$:
\begin{equation} \label{eq:zm}
z^m=\mathrm{u}(m)
\end{equation}
where $\mathrm{u}$ is a max pooling function and $z^m\in \mathbb{R}^{w\times h}$. Then, we can use $z^m$ to filter the non-defective parts of $z^x$ and to replace the corresponding parts in $y$:
\begin{equation} \label{eq:z}
z=z^y\odot(1 - z^m)+z^x\odot z^m
\end{equation}
where $z\in \mathbb{R}^{w\times h}$ is the latent feature. 

Note that the defect-free downsampling is an iterative process. We divide the process into a set of downsampling steps. The input will be transformed to the final discrete latent variable $z$ gradually. During the process, the defective region $m$ also evolves. Formally, we define the whole process as
\begin{align}
\begin{aligned}
h^x_i &= \mathrm{e}^{\mathrm{DF}}_i(h_{i-1}, m_{i-1}) \\
h^y_i &= \mathrm{e}_i(h_{i-1}) \\
h_{i} &= h^y_i\odot(1 - m_i)+h^x_i\odot m_i \\
m_i &= \mathrm{u}_i(m_{i-1}) \\
\end{aligned}
\end{align}
where $\mathrm{e}_i$ or $\mathrm{e}^{\mathrm{DF}}_i$ is a single step of normal or defect-free downsampling and $h_i$ is the result of the $i$-th downsampling step. The stride and kernel size of $\mathrm{u}_i$ is the same as the stride of the convolution in $\mathrm{e}_i$. Let $m_0 = m$ be the initial value of the process and $T$ be the total number of downsampling steps. In training, $h_0 = y$ but in inference, $h_0 = x$.
Thus, the final latent variable of the image is $z = h_T$. The corresponding masked tokens can be located by $z^m = m_T$.

With a learnable codebook $B$ trained by DF-VQGAN, the quantized tokens $\tilde{z}$ is:
\begin{equation} \label{eq:zi} 
\tilde{z}_i=\mathop{\arg}\mathop{\min}_{j}||z_i-B_j||^2,
\end{equation}
where $\tilde{z}\in \mathbb{R}^{w\times h}$ denotes the discrete tokens. By Eq.~(\ref{eq:zy})$\sim$(\ref{eq:zi}), the Encoder $q(z|y, x, m;\phi)$ can be modeled in a defect-free way, considering encoding only the non-defective regions in $x$ and use these non-defective regions to update the final $\tilde{z}$. 

\paragraph{Defect-free Upsampling} Following the VQGAN architecture, we first embed the latent variable $\tilde{z}$ by looking up the VQGAN dictionary via $\hat{h} = B[\tilde{z}]$, which is then fed into the upsampling function:

\begin{equation} \label{eq:y'}
y'=\mathrm{d}(\hat{h})
\end{equation}
where $\mathrm{d}$ is a upsampling function, $y'\in \mathbb{R}^{W\times H\times C}$. Different from the downsampling process in Eq.~(\ref{eq:zx}), we do not need to do defect-free operations, since there are no defective regions in $z$ because this region has been inpainted by either autoregressive model in inference or known in training.

We further use $m$ to filter the non-defective parts of z by,
\begin{equation} \label{eq:yhat}
\hat{y}=(1 - {m})\odot\frac{1}{\tau + 1}(y' +\tau x) + m\odot y'
\end{equation}

Correspondingly, upsampling is a reversed iterative process of downsampling. We propose \textit{\textbf{symmetrical connection}} to avoid the information loss of non-defective region. The hidden state of the non-defective region in the symmetrical downsampling step will be mixed with the output of the previous upsampling step:
\begin{align}
\begin{aligned}
\hat{h}_i &= \mathrm{d}_{i+1}(\hat{h}'_{i+1}) \\
\hat{h}'_i &= (1 - {m}_{i})\odot\frac{1}{\tau + 1}(\hat{h}_{i} +\tau h_{i}) + m_{i}\odot\hat{h}_{i}
\end{aligned}
\end{align}
where $d_i$ denotes a single step of upsampling and $\tau$ is the coefficient of information from the downsampling step. We set $\tau=\infty$ for $i=0$ due to we need remains non-defective region unchanged and $\tau=1$ for others. $d_i$ is the combination of a series of functions: attention, normalization, convolution, and nearest upsampling.

Symmetrically, the initial state of the downsampling is $\hat{h}_T = B[\tilde{z}]$ and the final output image is $\hat{x} = \hat{h}_0$ which is reconstructed based on latent variable $\tilde{z}$. During training, latent variable $\tilde{z}$ is copied from defect-free downsampling directly and in inference, latent variable $\tilde{z}$ was predicted by MP-S2S, which will be introduced in the following.

Following VQGAN, the training objective is:
\begin{align}
\begin{aligned}
\footnotesize
\mathcal{L}^{V} &= ||I-\hat{I}||_2^2+||\mathrm{sg}[\mathrm{E}(I)]-B[z]||_2^2\\
&+||\mathrm{E}(I)-\mathrm{sg}[B[z]]||_2^2, \\
\mathcal{L}^{P} &= ||\mathrm{Q}(I)-\mathrm{Q}(\hat{I})||_2^2,\\
\mathcal{L}^{G} &= \mathrm{log}\mathrm{D}(I)+\mathrm{log}(1-\mathrm{D}(\hat{I})).\\
\end{aligned}
\end{align}
where $\mathrm{E}$ and $\mathrm{D}$ respectively denote the defect-free downsampling and defect-free upsampling, and $\mathrm{Q}$ denotes a CNN-based module to obtain the conceptual representation of the image. The overall learning objective of DF-VQGAN is:
\begin{align}
\begin{aligned}
\mathcal{L}_1 &=\mathcal{L}^{V}+\mathcal{L}^{P}+\mathcal{L}^{G}
\label{loss_vae}
\end{aligned}
\end{align}
We use the full image $y$ with randomly generated mask metrics $m$ to train our DF-VQGAN.

\subsection{MP-S2S}\label{sec:mp}

\begin{figure}[h]
	\centering
	\includegraphics[width=2.5in]{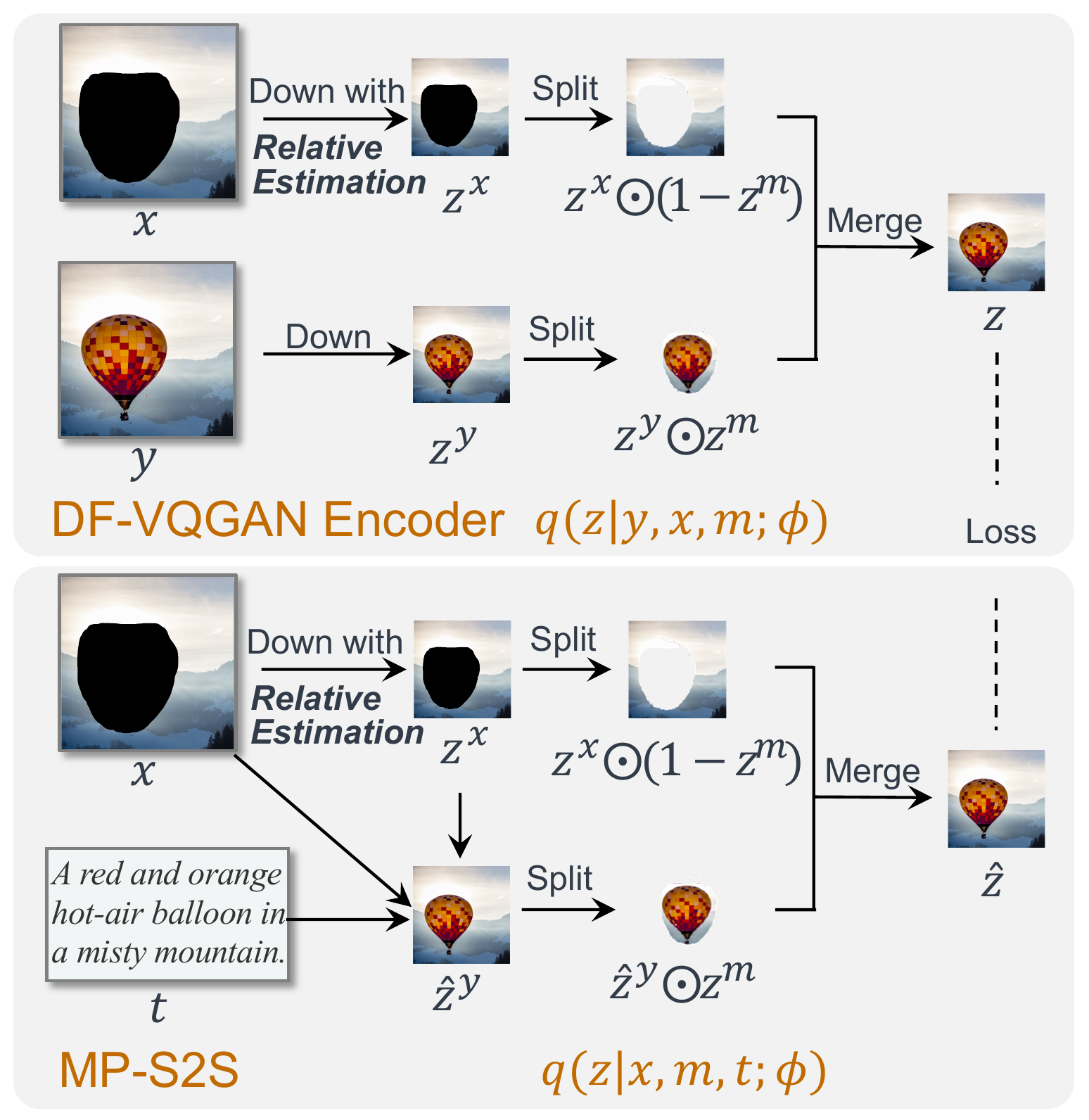}
	\caption{Illustration of our proposed MP-S2S during training.}
	\label{fig:mp}
\end{figure}

As noted in Sec.~\ref{sec: df}, we model the probabilistic density function $q(z|y,x,m;\phi)$ using the Kullback-Leibler divergence term $\mathbb{KL}_{z\sim q(z|y,x,m;\phi)}\left[q(z|y,x,m;\phi)||p(z|x, m, t;\theta)\right]$ by a DF-VQGAN encoder, while the key to solve this term relies on the modeling of density function $p(z|x, m, t;\theta)$. 
In this section, we propose a multi-perspective sequence to sequence (MP-S2S). As shown in Fig.~\ref{fig:mp}, MP-S2S encodes the information from three perspective including 1) the input text $t$, pixel-level defective image $x$ and token-level defective representation $z^x$, and decodes to predict the non-defective token-level representation $\hat{z}^y$.

\paragraph{Multimodal Encoding}

For Guidance Text from language modalities, following BERT~\cite{devlin2018bert}, we tokenize the text with BPE and embed them to the representation sequence $t$, where $t_i \in \mathbb{R}^D$ denotes the representation of each token and $D$ denotes the dimension of the representation. We encode token representation sequence $t$ to get its representation by $c^{\mathrm{t}} = \mathrm{E}^{\mathrm{t}}(t)$.

For defective image, the model treats it from two perspective, high-level VAE and low-level representation. For low-level vision, following ViT~\cite{dosovitskiy2020image}, we transform the image to a sequence of patches without overlapping $x^{\mathrm{p}} = (x_1^{\mathrm{p}}, x_2^{\mathrm{p}}, ...)$. Then, we directly encode the image patch sequence to get representation from low-level pixel perspective by $c^{\mathrm{l}} = \mathrm{E}^{\mathrm{l}}(x^{\mathrm{p}})$. For high-level representation, we embed the discrete token sequence $z$ with its Codebook $B$ in DF-VQGAN and encode them by $c^{\mathrm{h}} = \mathrm{E}^{\mathrm{h}}(B[z])$. Note that masked tokens, which can be located by $z^m$, will be replaced to a special trainable vector.
We use Transformer encoder as the architecture of $\mathrm{E}^{\mathrm{h}}$, $\mathrm{E}^{\mathrm{l}}$ and, $\mathrm{E}^{\mathrm{t}}$.

Thus, the integrated representation from two modalities can be formulated as $c = [c^{\mathrm{t}}; c^{\mathrm{l}}; c^{\mathrm{h}}]$.

\paragraph{Autoregressive Decoding}

The integrated representation $c$ from two modalities can be considered as the condition of the Decoder. The Decoder aims to predict the missing VAE tokens based on the condition and predicted tokens: $P(z_k|z_{<k}, c)$, where $z_k$ is the $k$-th token and $z_{<k}$ is token sequence before the $k$-th token. 


Following MASS \cite{song2019mass}, we only decode the masked token which can be located by $z^m$ in DF-VQGAN. The training objective of MP-S2S is:
\begin{equation}
    \mathcal{L}_2 = -\sum_{k}\mathrm{log}P(z_k|z_{<k}, c).
    \label{loss_ae}
\end{equation}

Combining Eq. (\ref{loss_vae}) and Eq. (\ref{loss_ae}), the final objective of NÜWA-LIP can be formulated by the sum of them:
\begin{equation}
    \mathcal{L} = \mathcal{L}_1 + \mathcal{L}_2
    \label{loss_all}.
\end{equation}
The training data are generated by pre-trained DF-VQGAN on image-text pairs. We use the full image $y$ with a randomly generated mask matrix $m$ to obtain the $z$ and $z^m$ from pre-trained DF-VQGAN.

During inference, we use MP-S2S to predict the masked tokens in $\tilde{z}$, which is obtained from defect-free downsampling. Then predicted tokens $\hat{z}^y$ will replaced the tokens in $\tilde{z}$ by $\hat{z} = \tilde{z}\odot(1 - {z^m}) + \hat{z}^y\odot{z^m}$. The $\hat{z}$ will be the input latent variable of defect-free upsampling in DF-VQGAN. As shown in Fig.~\ref{fig:model}, the inpainting result can be generated by such a pipeline.
\section{Experiments}

\subsection{Implementation Details}

For DF-VQGAN, we set the code-book size to $8,192$, the learning rate to $5\times 10^{-6}$, and the batch size to $200$.
We use ImageNet \cite{deng2009imagenet} as the pre-training corpus, where each image is resized to $256^2$.
For each image, the size of discrete visual tokens is set to $32^2$.

For MP-S2S, we set the number of layers in the encoder to $12$, the number of layers in the decoder to $24$, the hidden size to $1,024$, the learning rate to $5\times 10^{-4}$, and the batch size to $320$. 
The text encoder of MP-S2S is initialized with the text encoder of CLIP \cite{radford2021learning}. 
The patch size used in the low-level encoder is set to $16^2$.
We use Adam \cite{kingma2014adam} Optimizer with a warm-up ratio of $5\%$ and a dropout ratio of $10\%$. 
We use Conceptual Captions \cite{sharma2018conceptual} as the pre-training corpus.

The final pre-trained model has a total of $1.7$ billion parameters. We pre-trained the whole NÜWA-LIP with $64$ A100 GPUs for two weeks in total.

\subsection{Experiments Setup}

\paragraph{Evaluation Datasets}

\begin{table}[t]
\caption{Statistics of the evaluation datasets.}
\label{data}
\vskip 0.15in
\begin{center}
\begin{small}
\begin{sc}
\begin{tabular}{lcc}
\toprule
Dataset & Image-Text Pairs & Mask Ratio \\
\midrule
MaskCOCO & 5000 & 31.5\% \\
MaskFlickr & 1000 & 48.3\% \\
MaskVG & 10000 & 14.6\% \\
\bottomrule
\end{tabular}
\end{sc}
\end{small}
\end{center}
\vskip -0.1in
\end{table}

\label{datasets}

In order to evaluate the proposed model, we built up three evaluation datasets, MaskCOCO, MaskFlickr, and MaskVG based on MSCOCO \cite{lin2014microsoft}, Flickr \cite{young2014image} and VG \cite{krishna2017visual}, respectively. 
We follow Karpathy split \cite{karpathy2015deep} and select the test split of MSCOCO and Flickr to build MaskCOCO and MaskFlickr. 
For MaskVG, we sample 10,000 samples randomly from the VG dataset. 
For each image-text pair, the original image and corresponding caption are considered as full image and guidance text. 
Each image will be cropped and resized to the resolution of $256 \times 256$. The defective image is generated by masking either one bounding box of the object or a random irregular region in the original image. 
The statistics of the datasets are listed in Tab.~\ref{data}.

\paragraph{Evaluation Metric}

To evaluate the quality of the inpainted image under the language guidance, the FID score \cite{heusel2017gans} is chosen as the metric. 
We evaluate the representation similarity of concepts in the inpainted image and the original image. The lower FID score denotes the higher quality of the inpainted image.

\paragraph{Baselines}

We compare NÜWA-LIP with two strong baselines. GLIDE \cite{nichol2021glide} is an effective diffusion-based model for image generation and editing. We use the inpainting version of GLIDE in the official repository. NÜWA \cite{wu2021n} is another effective model for vision generation and editing. 
We re-implement NÜWA and re-train NÜWA following their description in the paper.
Additionally, we modified NÜWA to NÜWA-P by pasting the non-defective region on the inpainting result to make a comprehensive comparison.
Each model will output two images for each input image.  

\subsection{Overall Results}

\begin{table}[t]
\caption{Overall results of Language Guided Image Inpainting. We use FID (lower is better) as the evaluation metric.}
\label{overall}
\vskip 0.15in
\begin{center}
\begin{small}
\begin{sc}
\begin{tabular}{lccc}
\toprule
Model & MaskCOCO & MaskFlickr & MaskVG \\
\midrule
\makecell[l]{GLIDE} & 13.5 & 51.9 & 9.0 \\
\makecell[l]{NÜWA} & 21.4 & 59.5 & 18.5 \\
NÜWA-P & 20.6 & 54.2 & 17.7 \\
\makecell[l]{NÜWA-LIP} & \textbf{12.0} & \textbf{42.5} & \textbf{8.5} \\
\midrule
\makecell[l]{NÜWA-LIP\\(finetune)} & \textbf{10.5} & - & - \\
\bottomrule
\end{tabular}
\end{sc}
\end{small}
\end{center}
\vskip -0.1in
\end{table}

We first compare NÜWA-LIP with the two baseline models without fine-tuning.
Referring to Tab.~\ref{overall}, we can observe NÜWA-LIP achieves state-of-the-art performance on all datasets. 
Our proposed method outperforms the highest baseline GLIDE with 1.5 FID on MaskCOCO, 9.4 FID in MaskFlickr, and 0.5 on MaskVG, which shows the effectiveness of NÜWA-LIP. We suggest that the improvement is ascribed to that NÜWA-LIP can not only remain non-defective region unchanged but also avoid inaccurate or incomprehensive encoding.

We then fine-tune  NÜWA-LIP on MSCOCO datasets as an additional model.
The fine-tuned performance obtains FID of 10.5 on MaskCOCO, which shows in-domain fine-tuning can further improve the ability of NÜWA-LIP.

\subsection{Human Evaluation}

\begin{figure}
	\centering
	\subfigure{
	\includegraphics[width=2.9in]{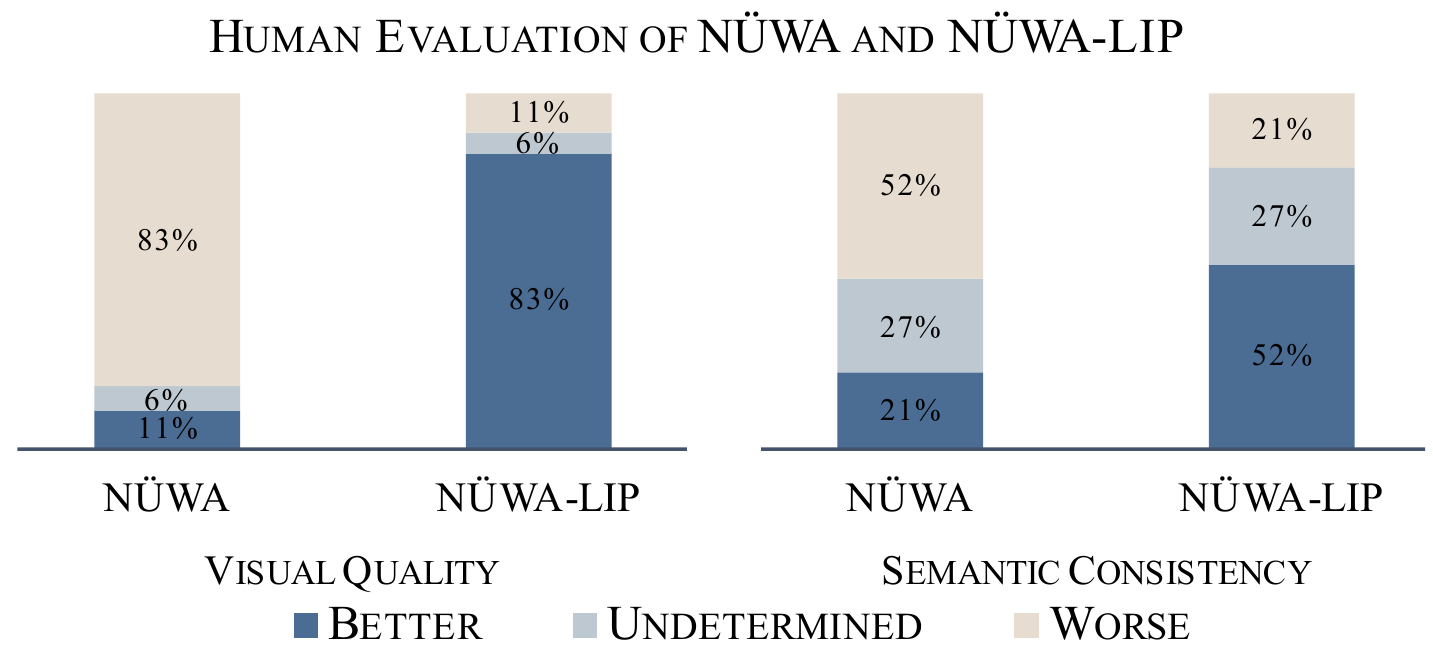}
	}
	\subfigure{
	\includegraphics[width=2.9in]{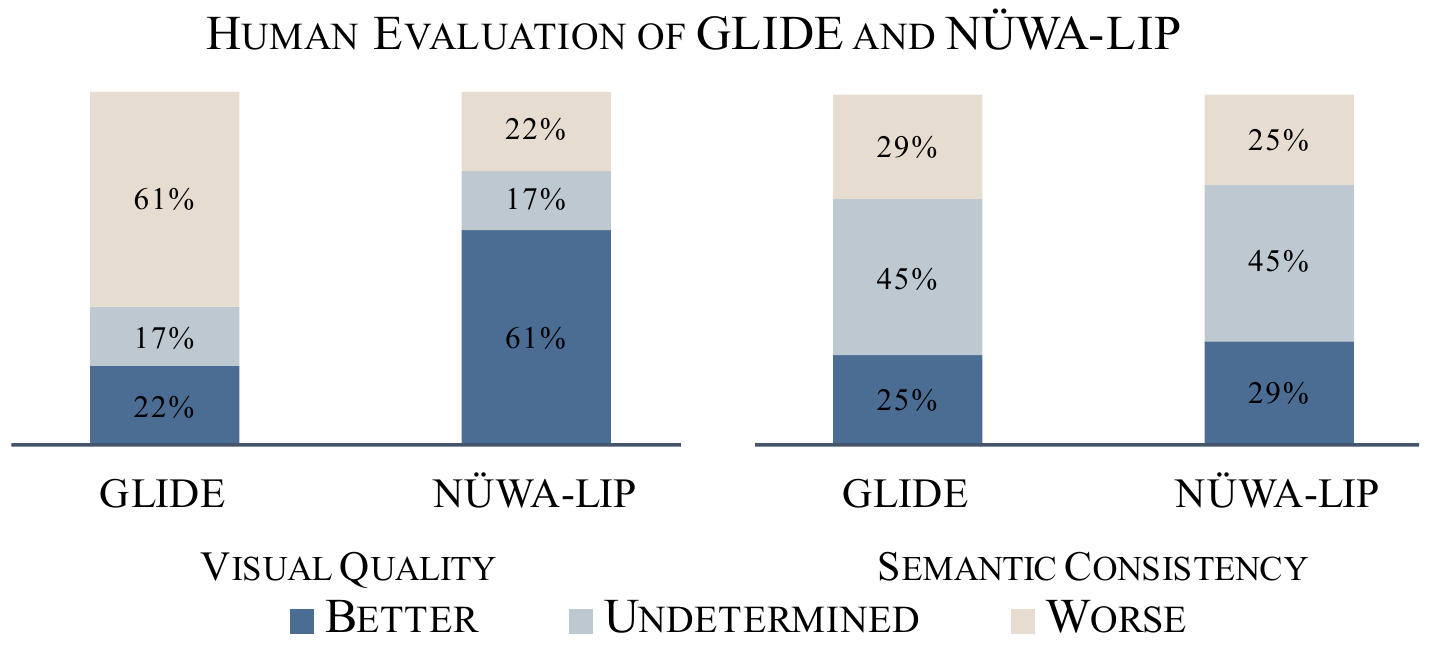}
	}
	\caption{Human Evaluation of NÜWA-LIP and baselines.}
	\label{fig:human}
\end{figure}
To further compare the quality of NÜWA-LIP and baselines, we evaluate the inpainting results by humans. We randomly sample 500 samples from the MaskCOCO dataset and ask human annotators to compare the visual quality and semantic consistency of NÜWA-LIP, NÜWA, and GLIDE.
The visual quality represents the quality of entirety, including the transition of the edge or continuity of pixels. The semantic consistency represents the compliance of the language guidance.
Referring to Fig.~\ref{fig:human}, NÜWA-LIP's visual quality increased significantly in comparison with NÜWA or GLIDE. We also achieve better semantic consistency.

\subsection{Effectiveness of DF-VQGAN}

\label{inp}

\begin{figure}
	\centering
	\includegraphics[width=2.7in]{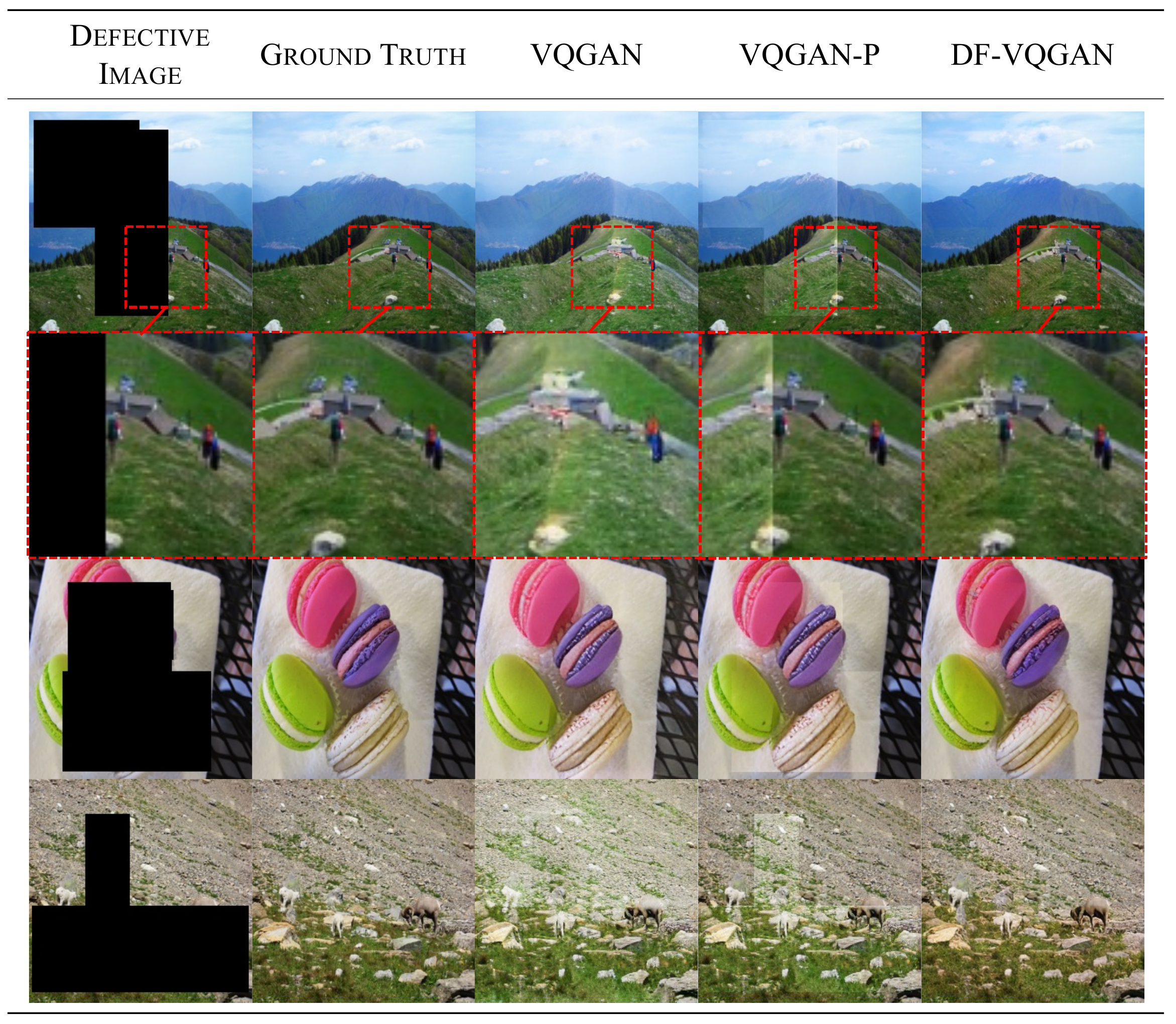}
	\caption{Illustration on Oracle Image Inpainting.}
	\label{fig:case_dfvqgan}
\end{figure}

We first compare DF-VQGAN with original VQGAN on the \textbf{image reconstruction} task (Tab.~\ref{df-vqgan}). whose goal is to transform each complete image into discrete latent tokens and then reconstruct the image based on these discrete tokens. This task can assess the image encoding and decoding  performance of VQGAN, which are critical to NÜWA-LIP.

From Tab.~\ref{df-vqgan}, one can see that DF-VQGAN performs better than VQGAN in the same setting (\emph{i.e.}, resolution and vocab size). This improvement may be ascribed to that the training of DF-VQGAN covers both complete image reconstruction and incomplete image reconstruction illustrated in Fig.~\ref{fig:df}, which makes the resulting model more robust.



We then compare DF-VQGAN and original VQGAN on the \textbf{oracle image inpainting} task (Tab.~\ref{df-vqgan2}), whose goal is to reconstruct the image given the ground-truth discrete tokens of the defective region (predicted based on the ground-truth image) and the discrete tokens of the rest region (predicted based on the defective image). 

From Tab.~\ref{df-vqgan2}, one can see that DF-VQGAN performs better than VQGAN in the same setting (\emph{i.e.}, resolution and vocab size), which verifies the effectiveness of DF-VQGAN for the image inpainting task. Fig.~\ref{fig:case_dfvqgan} shows some visualization examples of VQGAN and DF-VQGAN on this task. 



In Fig.~\ref{fig:case_dfvqgan}, we compare DF-VQGAN and VQGAN with VQGAN-P, which copies the non-defective region from the input image and uses it to substitute the reconstruction part of the same region by VQGAN. 
In comparison to VQGAN-P, DF-VQGAN has a significantly better transition of the non-defective region and inpainted region.



\begin{table}[t]
\caption{Image Reconstruction results on ImageNet.}
\label{df-vqgan}
\vskip 0.15in
\begin{center}
\begin{small}
\begin{sc}
\begin{tabular}{lccc}
\toprule
Model & Resolution & Vocab Size & \hspace*{3pt}FID$^{\downarrow}$ \\
\midrule
VQGAN & $256^2 \to 32^2$ & 8192 & 1.47 \\
DF-VQGAN & $256^2 \to 32^2$ & 8192 & \textbf{1.38} \\
\midrule
VQGAN & $256^2 \to 16^2$ & 12288 & 5.48 \\
DF-VQGAN & $256^2 \to 16^2$ & 12288 & \textbf{5.16} \\
\bottomrule
\end{tabular}
\end{sc}
\end{small}
\end{center}
\vskip -0.1in
\end{table}

\begin{table}[t]
\caption{Oracle Image Inpainting results on ImageNet.}
\label{df-vqgan2}
\vskip 0.15in
\begin{center}
\begin{small}
\begin{sc}
\begin{tabular}{lccc}
\toprule
Model & Resolution & Vocab Size & \hspace*{3pt}FID$^{\downarrow}$ \\
\midrule
VQGAN & $256^2 \to 32^2$ & 8192 & 2.04 \\
DF-VQGAN & $256^2 \to 32^2$ & 8192 & \textbf{0.80} \\
\midrule
VQGAN & $256^2 \to 16^2$ & 12288 & 7.15 \\
DF-VQGAN & $256^2 \to 16^2$ & 12288 & \textbf{2.95} \\
\bottomrule
\end{tabular}
\end{sc}
\end{small}
\end{center}
\vskip -0.1in
\end{table}

\subsection{Ablation Studies}


%
%

\paragraph{DF-VQGAN}

The ablation of DF-VQGAN aims to verify whether the symmetrical connection and relative estimation play the expected roles. \textsc{DF-VQGAN/sr} and \textsc{DF-VQGAN/s} respectively denote the model without both symmetrical connection and relative estimation or symmetrical connection.
We re-train DF-VQGAN under different settings and evaluate Oracle Image Inpainting mentioned in Sec~\ref{inp} on ImageNet. Referring to Tab.~\ref{ablation-dfvqgan}, we can observe that relative estimation reduces the FID by 2.1 and the symmetrical connection further provides a decrease of 0.3.

\begin{table}[t]
\caption{Ablation results of DF-VQGAN on MaskCOCO.}
\label{ablation-dfvqgan}
\vskip 0.15in
\begin{center}
\begin{small}
\begin{sc}
\begin{tabular}{lccc}
\toprule
\multirow{2}{*}[-1.5pt]{Model} & \multicolumn{2}{c}{Component} & \multirow{2}{*}[-1.5pt]{\hspace*{3pt}FID$^{\downarrow}$}\\
\cmidrule{2-3} & Sym.Con & Rel.Est  \\
\midrule
DF-VQGAN/sr & $\times$ & $\times$ & 14.9 \\
DF-VQGAN/s & $\times$ & $\surd$ & 12.8 \\
\midrule
DF-VQGAN  & $\surd$ & $\surd$ & \textbf{12.5} \\
\bottomrule
\end{tabular}
\end{sc}
\end{small}
\end{center}
\vskip -0.1in
\end{table}

\paragraph{MP-S2S} To verify whether the multi-perspective benefits Language Guided Image Inpainting, we conduct the ablation on MP-S2S by eliminating one or two of its perspectives. \textsc{MP-S2S/L}, \textsc{MP-S2S/H}, or \textsc{MP-S2S/T} denotes the MP-S2S model without low-level pixel encoder, high-level token encoder or text encoder. \textsc{MP-S2S/HL} denotes the MP-S2S model which only has a text encoder.
To exclude the influence from other components, we use the same original VQGAN as a backbone. We re-train the MP-S2S under different settings on MSCOCO and evaluate the performance of Language Guided Image Inpainting results on MaskCOCO. Referring to Tab.~\ref{ablation-mps2s}, one can observe the text perspective provide the highest gain of 8.5 and high-level tokens and low-level pixels perspective gains the model with FID of 1.2 and 0.6, respectively.

\begin{table}[t]
\caption{Ablation results of MP-S2S on MaskCOCO.}
\label{ablation-mps2s}
\vskip 0.15in
\begin{center}
\begin{small}
\begin{sc}
\begin{tabular}{lcccc}
\toprule
\multirow{2}{*}[-1.5pt]{Model} & \multicolumn{3}{c}{Component} & \multirow{2}{*}[-1.5pt]{\hspace*{3pt}FID$^{\downarrow}$}\\
\cmidrule{2-4} & Text & High & Low \\
\midrule
MP-S2S/hl & $\surd$ &$\times$ & $\times$ & 29.4 \\
MP-S2S/l & $\surd$ &$\surd$ & $\times$ & 27.4 \\
MP-S2S/h & $\surd$ & $\times$ & $\surd$ & 26.8 \\
MP-S2S/t & $\times$ & $\surd$ & $\surd$ & 34.7 \\
\midrule
MP-S2S & $\surd$ & $\surd$ & $\surd$ & \textbf{26.2} \\
\bottomrule
\end{tabular}
\end{sc}
\end{small}
\end{center}
\vskip -0.1in
\end{table}

\subsection{Case Studies}

\begin{figure*}
	\centering
	\includegraphics[width=17cm]{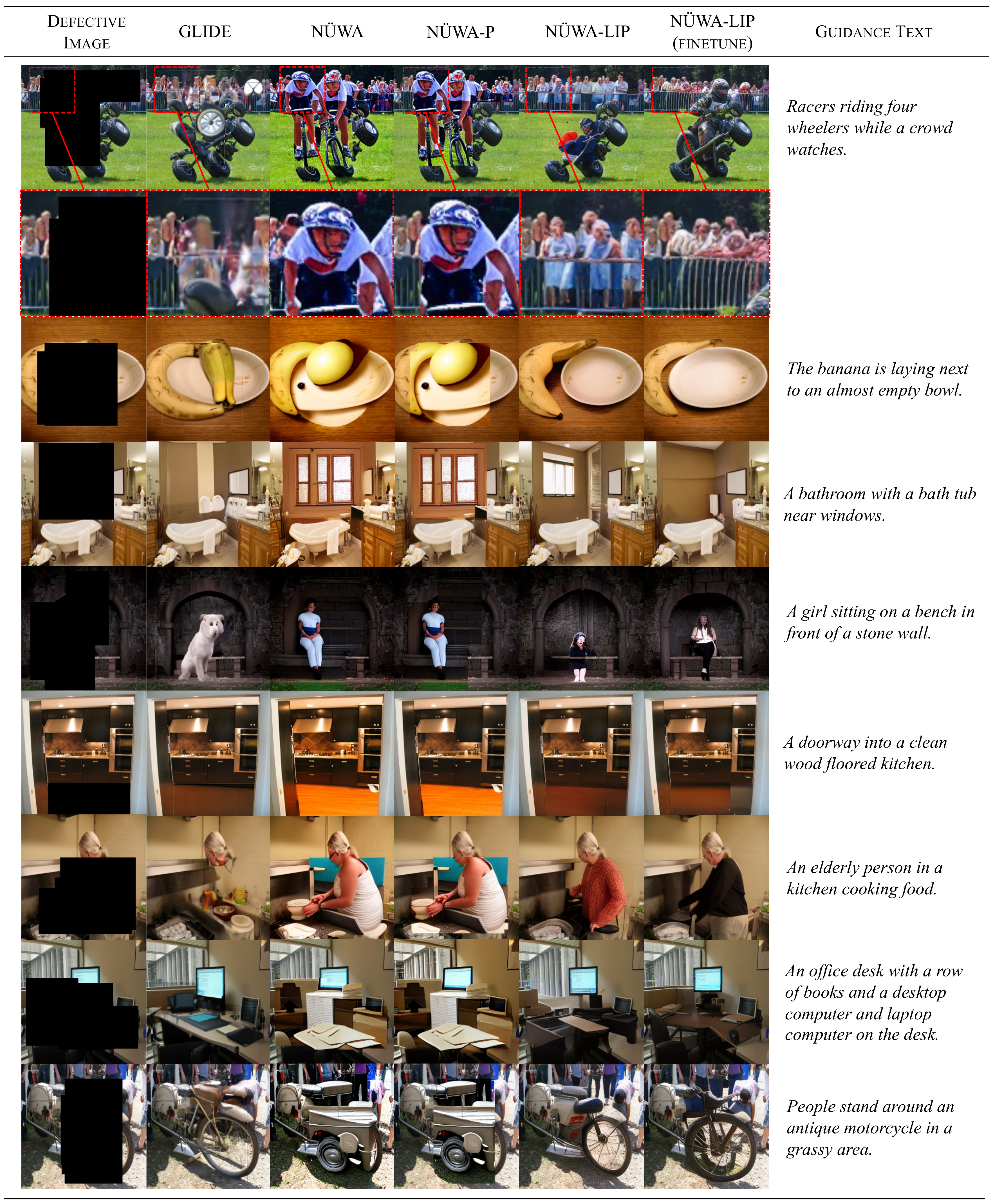}
	\caption{Language Guided Image Inpainting results of different models.}
	\label{case}
\end{figure*}

To better reveal the quality of NÜWA-LIP, we selected several cases to show the inpainting results generated by NÜWA-LIP and baselines. Referring to Fig.~\ref{case}, one can observe that (1) compared with NÜWA, NÜWA-LIP remains the correct hue of the whole image, (2) compared with baselines, NÜWA-LIP makes a better transition of non-defective and inpainted region, (3) NÜWA-LIP generate more details which accord with the guidance text, and (4) fine-tuning NÜWA-LIP further improves the quality of the inpainting.
\section{Conclusion}

In this paper, we proposed NÜWA-LIP for Language Guided Image Inpainting. Our contributions are three-fold: (1) We proposed DF-VQGAN, which can control receptive spreading protect information. (2) We proposed MP-S2S, which further enhances visual information from complementary perspectives. (3) We built up 3 open-domain benchmarks, where NÜWA-LIP is also superior to baselines.


\nocite{langley00}
\bibliography{example_paper}
\bibliographystyle{icml2022}



\end{document}